\documentclass[manuscript,screen]{acmart}
\usepackage[ruled,vlined]{algorithm2e}
\AtBeginDocument{%
  \providecommand\BibTeX{{%
    \normalfont B\kern-0.5em{\scshape i\kern-0.25em b}\kern-0.8em\TeX}}}

\setcopyright{acmcopyright}
\copyrightyear{2022}
\acmYear{2022}
\acmDOI{10.1145/1122445.1122456}

\acmBooktitle{ACM Transactions on Asian and Low-Resource Language Information Processing}
\acmPrice{15.00}
\acmISBN{2375-4699}

\begin{document}

\title{Evaluating Persian Tokenizers}

\author{Danial Kamali}
\email{danial_kamali@comp.iust.ac.ir}
\orcid{0002-2652-2339}
\affiliation{%
  \institution{Iran University of Science and Technology}
  \streetaddress{Narmak}
  \city{Tehran}
  \state{Tehran}
  \country{Iran}
  \postcode{16846-13114}
}
\author{Behrooz Janfada}
\email{behrooz.janfada@gmail.com}
\orcid{0001-9370-010X}
\affiliation{
  \institution{Iran University of Science and Technology}
  \streetaddress{Narmak}
  \city{Tehran}
  \state{Tehran}
  \country{Iran}
  \postcode{16846-13114}
}
\author{Mohammad Ebrahim Shenasa}
\email{me.shenasa@iau-tnb.ac.ir}
\orcid{0002-0873-424X}
\affiliation{%
  \institution{Islamic Azad University Tehran North Branch}
  \city{Tehran}
  \state{Tehran}
  \country{Iran}
}

\author{Behrouz Minaei Bidgoli}
\email{b_minaei@iust.ac.ir}
\orcid{0002-9327-7345}
\affiliation{%
  \institution{Iran University of Science and Technology}
  \streetaddress{Narmak}
  \city{Tehran}
  \state{Tehran}
  \country{Iran}
  \postcode{16846-13114}
}

\renewcommand{\shortauthors}{Kamali, et al.}

\begin{abstract}
  Tokenization plays a significant role in the process of lexical analysis. Tokens become the input for other natural language processing tasks, like semantic parsing and language modeling. Natural Language Processing in Persian is challenging due to Persian's exceptional cases, such as half-spaces. Thus, it is crucial to have a precise tokenizer for Persian.
     This article provides a novel work by introducing the most widely used tokenizers for Persian and comparing and evaluating their performance on Persian texts using a simple algorithm with a pre-tagged Persian dependency dataset. After evaluating tokenizers with the F1-Score, the hybrid version of the Farsi Verb and Hazm with bounded morphemes fixing showed the best performance with an F1 score of 98.97\%.
\end{abstract}


\maketitle

\section{Introduction}
In natural language processing, tokenization is the process of converting a string of characters to a series of tokens. Every stream of characters must turn into a series of distinct meaningful units named tokens before any further process. Until now, various tokenizers have been introduced for Persian\footnote{In this article, Persian represents the Persian language}. Some are standalone tools, and others are part of more extensive natural language processing tools. However, so far, no comparative evaluation has been provided to measure the performance of these tokenizers. This research introduces the most widely used and well-known Persian tokenizers and evaluates their performance for Persian texts on a Persian dependency dataset.
	
     In section 2 of this article, we review previous works that have been done in this area. Then in section 3, we define tokenization. We also discuss its main problems in Persian. Section 4 introduces the most well-known Persian tokenizers and their methods. Then, in section 5, we introduce the dataset and our evaluation method. In section 6, we present the results of the evaluations. Finally, in the last section, based on the results of the experiments, we present the conclusion of this research and discuss the feasibility of future work.
    The following are
the key contributions of our work:
\begin{enumerate}
    \item introducing widely used tokenization tools
    \item proposing an improved version of the current best tokenizer
    \item proposing an evaluation method and evaluating tokenizers using commonly used metrics
\end{enumerate}

	\section{Related Work}
	Tokenization has always been considered one of the essential parts of the text processing pipeline. In a study by Webster and Jonathan ~\cite{Webster}, they point out the complexities and importance of tokenization. In two consecutive studies ~\cite{McNamee2003} ~\cite{McNamee2004}, McNamee and Mayfield used n-grams for tokenization in European languages and introduced this method as a "relatively fast" method for text preprocessing. In a study by Grefenstette and Tapanainen ~\cite{Grefenstette1994}, they mentioned tokenization as one of the essential preprocessing operations of the text. They addressed the problems and difficulties of this operation in English. Habash and Rambow ~\cite{Habash2005} propose a morphological method to tokenize words. 
    In recent years, numerous research has been done on the tokenization of Persian texts. There is also research on language-independent word tokenization methods. For example, Moreau and Vogel ~\cite{moreau-vogel-2018-multilingual} have introduced a method for training a multi-lingual tokenizer. They introduced a method to train a tokenizer using the Universal Dependency2 dataset. The result of their study shows that this method has achieved significant results in most languages. Also, this method could accomplish 100\% recall in Persian, but due to the authors' unfamiliarity with Persian, they could not correctly calculate the accuracy~\cite{moreau-vogel-2018-multilingual}.
    Right now, two of the state-of-the-art tokenization methods are Trankit ~\cite{vannguyen2021trankit}, and Stanza ~\cite{qi2020stanza}. These are multi-lingual deep learning based tools that support the Persian language in their tokenizers but do not offer Persian multi-word tokenization. Also, They are the best tokenization methods on the Universal Dependency datasets.

    The simplest way to tokenize Persian text is to separate the tokens based on the " "(space) character. However, using this method alone is not accurate enough. Therefore, in recent years, efforts have been made to improve the results of Persian tokenization. For example, Shamsfard et al. ~\cite{inproceedings} proposed a set of natural language preprocessing tools called Step-1 that performed better quality tokenization. Step-1 got 87\% accuracy on a 400 words dataset with 115 errors.
    After that, Sarabi et al. ~\cite{Sarabi2013ParsiPardazPL} introduced a new toolkit called ParsiPardaz, which adds new features such as normalization and spell-checking to Step-1. ParsiPardaz has reported 95\% precision on the Dadehgan dataset.
   Sobhe has provided another open-source toolkit called HAZM, one of the most well-known packages for preprocessing Persian texts. HAZM involves many preprocessing tools such as normalizer, tokenizer, and part of speech labeler. Also, Farsi Verb tokenizer was found by the author of this article while searching in the collection of Github codes ~\cite{fvt}.

    In 2018 ~\cite{ParsiVar} presented a new set of tools called ParsiVar, which includes tools needed for numerous Persian processing tasks. In their study, experiments have shown that HAZM is faster than Parsipardaz and Step-1, but it has worse accuracy than the others. ParsiVar normalizes the input text to standard, corrects the spacing among letters or words, marks words and sentences, extracts the root of the word, and finally performs shallow parsing on sentences. In their research ~\cite{ParsiVar}, two different solutions have been proposed to correct the distance problem. The first solution is based on predefined rules, and the second is based on machine learning. But in the end, the results of experiments in their study show that the ParsiVar achieves better results than previous tools. In 2019 \cite{8734954} Parandeh and Ghanbari investigated the N-gram language model for space correction, which resulted in an 81.94\% F1-score. In 2020 Mohammadi and Nassajian ~\cite{doostmohammadi-etal-2020-joint} specifically worked on the problem of half-space (zero-width non-joiner) in word segmentation. They approach these problems as a sequence labeling problem and show the significance of the half-space problem. Also, Mohammadi et al. ~\cite{ParsBERT} created the first BERT ~\cite{devlin2019bert} based text-processing tools in Persian, which has reached state-of-the-art results in various NLP tasks. 
    
	\section{Tokenization}
    Tokenization is separating meaningful units as tokens contained in a text character string. The resulting tokens are entered as input to the language processing systems. There are problems while performing this operation in Persian, which we will elaborate on here.
    	
	\subsection{Different Encoding}
	
In order to store characters in computer systems, characters must be encoded. There are many different ways to encode characters, such as UTF-8 and Unicode. This variety of encoding can confuse tokenizers. For example, a character encoded using the Unicode method may not be readable or get misrepresented on other systems. This problem can be solved by using all the different encodings in the tokenizer or using an adaptation method. In the adaptation, a preprocessing step must be performed before tokenization. All of the encodings will be converted to an appropriate encoding for the tokenizer during this preprocessing step. After the tokenization, the output gets converted to the original encoding.
	\subsection{Word}
	There are three types of compound words in Persian. The first type is compound words whose components are entirely separated by a space character. The second type consists of compound words whose components are separated by a half-space, and the third type consists of compound words whose members are connected.

	In other words, determining word and phrase boundaries is a complicated task in Persian. The challenge in Persian is that space cannot be considered a valid delimiter in many cases.

	\section{Tokenizers}
From all the aforementioned Persian tokenizers, some of them are publicly accessible. In this section, we introduce them and summarize their method of functionality.

	\subsection{ParsiVar}
	This tokenizer uses these solutions to solve the boundary problem. It has shown an 89.5\% F-1 score on the Bijankhan dataset in the tokenization task.
	
	\begin{itemize}
		\item Rule-based space correction:
		\begin{itemize}
			\item Some specific rules have been defined in the first step using regular expressions.

			\item There are still some words that do not match in any of the rules. These kinds of words usually consist of more than one part that we cannot extract a general rule for them, such as "goft o goo"
			(Conversation).To overcome this problem, they constructed a dictionary containing such words and checked their existence in the sentences.
		\end{itemize}
		
		\item Space correction based on learning :
		\begin{itemize}
			\item training set using 90\% of Bijan khan corpus ~\cite{Bijankhan2011}.
			\item tagged the multi token words using IOB tagging 
			format. 96.5\% of F1-score.   
		\end{itemize}
		
	\end{itemize}

	\subsection{Hazm}
This tool uses Persian word files, verb files, suffixes, and prefixes of verbs. The text cleaning operation replaces items such as links, emojis, emails, hashtags, and numbers with the correct formats, then divides the sentence according to the " " (space) character.
	\subsection{ParsBert}
This toolkit was published in 2020 ~\cite{ParsBERT}, and it is a mono-lingual BERT-based tool in Persian. ParsBERT supports subword tokenization which can eventually improve its performance. It uses the WordPiece, trained on the ParsBert pre-training dataset, ~\cite{kudo-2018-subword} method to achieve this goal.
	\subsection{SetPer}
	
This is a regex-based word tokenizer using PERL made by Mojgan Seraj based on her thesis \cite{inproceedings} that uses the Uplug ~\cite{Tiedemann2003} modular software platform, a system designed for the integration of text processing tools. The Uplug sentence segmenter and tokenizer is a rule-based program that can adapt to various languages by using regular expressions for matching common word and sentence boundaries. SetPer considers the full stop, the question mark, and the exclamation mark as sentence boundaries. Token separators in SetPer are apostrophe, brackets, colon, semicolon, dash, exclamation mark, question mark, etc. This tokenizer also handles numerical expressions, web URLs, abbreviations, acronyms, and titles.

\subsection{Farsi Verb}
This tokenizer ~\cite{fvt} is a regex-based tokenizer that includes several resources like the past tense set, the present tense set, and a set of transformations (Persian verb patterns). This tokenizer replaces the founded pattern with a corrected version. It searches for all possible combinations of verbs in each sentence. For example, suppose the phrase "har rooz az anha poul khasteh nemidosheh ast." as the input to this algorithm. The algorithm first examines all the transformations of the Persian verb in the sentence. It checks all combinations related to the past tense or present tense, with appropriate prefixes or suffixes. In this example, the pattern "past tense(khast)+eh+nemishode+ast" is found, then in the boundary where these patterns are found, it replaces the letter " " (space) with "\_" (under-line) to form the whole phrase into a single token, and finally, "har rooz az anha poul khast\_eh\_nemidosheh\_ast." will result as the output.

	\section{Experiments}
	This section first elaborates on our evaluation-set creation method and describes the primary dataset. Second, it introduces an evaluation method and explains the used notations.

	\begin{table}
			\caption{Dataset Properties.}
			\label{tab:ds_properties}
			\begin{tabular}{cl} 
    \toprule
    Property & Value\\
    \midrule
    Number of Sentences & 29,982 \\ 
                Average Sentence Length &16.61 \\ 
                Number of Distinct Words &37,618 \\
                Number of Distinct Lemmas &22,064 \\ 
                Number of Verbs &60,579 \\ 
                Number of Verb Lemmas &4,782 \\ 
                Average Frequency of Verb Lemmas &12.67\\ 
                Number of Tokens & 399,152 \\
				Max Token Length & 39 \\ 
				Average Token Length & 10 \\ 
				  \bottomrule
			\end{tabular}
	\end{table}
	
	\subsection{Data}
	
We used the training section of Persian Dependency\cite{dataset}. Then we replaced every half-space with a full-space. Then we split every sentence by space into different lines so that every line of the new data consisted of an individual word that may not be correctly tokenized anymore. The primary dataset properties and statistics are shown in Table \ref{tab:ds_properties}.

	\begin{algorithm}
	  \caption{Evaluator's pseudocode}

		\KwData{test\_data, tokenizer\_result }
		\KwResult{\#Errors, F1-score, Accuarcy }
		initialization\;
		error = 0\;
		tp = 0\;
		fp = 0\;
		fn = 0\;
		\While{not at end of  test\_data}{
			read test\_data\_line, tokenizer\_result\_line \;
			
			\uIf{read test\_data\_line == tokenizer\_result\_line}
			{
		
				tp ++ \;
				continue \;
			}
			\uElseIf{is tokenizer\_line is wrong  }
			{
				error ++ \;
			}
			
			\eIf
			{
				read test\_data\_line\_length   $<$  tokenizer\_result\_line\_length\;    
			}
			{
				test\_data\_line = test\_data\_line + read test\_data\_line\;    
				fp ++ \;
			}
			{
				tokenizer\_result\_line = tokenizer\_result\_line + \\  read tokenizer\_result\_line\;    
				fn ++ \;    
			}
			
		}
	\label{algorithm1}
	\end{algorithm}
	The following statement can be helpful to clarify the algorithm. 
	In the comparison of the tokenizers' suggested token and actual tokens:
	\begin{enumerate}
	    \item if the actual and suggested units are the same, then it is a True Positive (TP).
	    \item if the tokenizer over-splits an actual token, then it is a False Positive (FP).
	    \item if the tokenizer under-splits an actual token, then it is a False Negative (FN).
	\end{enumerate}

	\subsection{Improvement by Bounded Morphemes Fixing  }
	After analyzing the errors of the aforementioned tokenizers, words bound morphemes were the most common problem that had not been fixed sufficiently in their implementation. To enhance the results, we checked the existence of a list of word prefixes and suffixes in the sentences and connected them to the before or afterward word. The suffix list includes plural signs and possessive signs, and this prefix list includes Persian word prefixes like "ba" and "na". The python implementation of this improvement on Farsi Verb Tokenizer is available at github\footnote{https://github.com/iamdanialkamali/Persian-Tokenizer}.

	\subsection{Evaluation}
To evaluate the results of tokenizers, we propose a simple algorithm shown in Algorithm 1. In the evaluation process, the comparative criteria of F1 score (Equation \ref{f1}) and accuracy (Equation \ref{acc}) have been used as primary metrics. The primitive criteria used here include TP (True Positive), FP (False Positive), FN (False Negative), and Error. The number of errors indicates the total number of tokenization errors. The number of TPs indicates the number of tokens that the system has correctly identified according to the labeled data. The number of FPs demonstrates the number of tokens that should not have been identified as a token compared to the labeled data but were incorrectly identified by the machine as a token. The number of FNs is the number of units that the machine should have identified as a token. However, the system has failed to detect them.

	\begin{equation}\label{pr}
	    Precision=\frac{TP}{TP+FP}
	\end{equation}
    \begin{equation}\label{rec}
    Recall=\frac {TP}{TP+FN}
    \end{equation}
    
	\begin{equation}\label{acc}
	Accuracy = \frac{TN+TP}{TP+TN+FN+FP}
	\end{equation}
	
	\begin{equation}\label{f1}
	F1 = \frac{2 * precision * recall}{precision + recall}
	\end{equation}

	\section{Results}

	        Contemplating the results in Table ~\ref{tk_errors}, one thing that is worths mentioning is that despite the high performance of state-of-the-art multi-lingual tools like Stanza and Trankit on the European languages, they can not keep up their performance in Persian. They fixed some errors in the text but added more errors in the process. The same goes for the SetPer tokenizer; analysis of its errors shows that it fixed a few mistakes but added more errors in the process. As shown in Table ~\ref{tk_errors}, the number of Hazm tokenizer's errors is far less than other tokenizers, and it has about 1.12\% better F1-score than ParsiVar, which is the second-best one. To better understand this difference, it is better to refer to the difference in the number of errors of these two tokenizers, which is about 20\% in error fixation, which is about 8000 errors. Hazm algorithm has performed better than other tools in F1-score, precision (equation \ref{pr}) and accuracy (equation \ref{acc}), and only in the recall metric, it has not achieved the best result. However, it is necessary to notice that Hazm has the best recall among the top-3 tokenizers.\\
	        As we can see in Table
                        
	        \begin{table}
	        \begin{minipage}{\linewidth}
		\begin{center}
			\caption{Tokenizers results on 399152 tokens.}
			\label{tk_errors}

			    \begin{tabular}{clllllll} 
			        \toprule
    				Tokenizer & \#Errors& Errors Fixed (\%) & Precision(\%) & Recall(\%)& F1(\%) & Accuracy(\%) & Time(s) \\
    				    \midrule
                    Stanza  &41922 & -0.06 &89.50 &\textbf{100} &94.46 &89.50 & 370 \footnote{on Tesla K80 GPU} \\
                    Trankit  &41828 & -0.03 &89.51 &\textbf{100} &94.47 &89.51 & 3681 $^a$ \\
                    
                    SetPer  &41800 & -0.03 &89.53 &\textbf{100} &94.47 &89.53 & \textbf{0.10} \\
                    Space Delimiter (Baseline)  &41669 & 0.00 &89.56 & \textbf{100} &94.50 &89.56 & 2.14 \\
    				Parsivar Without Normalization &41669 & 0.00 &89.56  &\textbf{100}  &94.50  &89.56 &  0.22\\
    				Bound Morphemes &33161 & 20.42 &91.64  &99.98  &95.63  &91.63 & 1.28 \\
    				
    				Hazm Without Normalization &31080 & 25.41 &92.21  &99.99  &95.94  &92.20 & 0.97 \\
    				
    				FarsiVerb  &22301& 46.48 & 94.48 & 99.92 &  97.12&94.40 & 1274 \\
    				Parsivar &18122 & 56.50 &97.05  &98.23  &97.634 &95.38 & 22.80 \\
    				Hazm &\textbf{9787}&  \textbf{76.51}& \textbf{97.57} &99.97 &\textbf{98.75} &\textbf{97.54} & 3.47 \\
    
    				    \bottomrule
    			\end{tabular}
    			 
		\end{center}
		\end{minipage}
	\end{table}
 
	        \ref{tk_normaliztion}, any of these tokenizers have different features and approaches that have their own advantages and disadvantages. The mix and matching approach is beneficial to benefit from the advantages of more than one of them and decrease the severity of their weaknesses. Table \ref{tk_errors_1} shows that mixing different tokenizers can result in better tokenization. In all of the tokenization experiments, the hybrid version of Hazm, Bound Morphemes, and FariVerb showed about 4\% improvement in the fixed errors. These tests were performed on Intel 8550u CPU and Tesla K80 GPU. Hazm tokenizer has shown the best run-time performance in this environment among all tokenizers with a positive error fixation rate. Farsi verb tokenizer, with concentration solely on verbs, offers a great accuracy on verbs, not on complex words, so it should only be used as an additional step for other tokenizers.\\
	        After analyzing the Hazm errors, despite the existence of verb parts joining and word affix fixing in its normalization process, there were still some problems in these areas. So, we tried to reduce these errors by mixing them with tokenizers with more focus in the respecting areas. It resulted in a 4\% increase in error fixation. After further analysis of the mistakes, it could be seen that most of the tokenization errors were because of irregular complex words. The improvement in this area needs gathering an extensive dataset of irregular complex words.

	\begin{table}
		\begin{center}
			\caption{Tokenizers Normalization Methods.}
			\label{tk_normaliztion}
			\begin{tabular}{ccccccc} 
		    \toprule
			Method &SetPer & FarsiVerb & Hazm&ParsiVar\\
			    \midrule
            Verb Parts Join   & & \checkmark  & \checkmark & \checkmark \\
			Word Affix Fixing      & \checkmark&   & \checkmark & \checkmark  \\
			Punctuation Spacing & & &\checkmark &   \\
			Statistical Space Correction & & &  & \checkmark \\
			Regex-Based Space Correction &\checkmark &\checkmark & \checkmark & \checkmark \\
			    \bottomrule
			\end{tabular}
		\end{center}
	\end{table}

 	\begin{table}
		\begin{minipage}{\linewidth}
		\begin{center}
			\caption{Tokenizers results for hybrid versions.}
			\label{tk_errors_1}
			
			\begin{tabular}{clllllll}
			    \toprule
				Tokenizer & \#Errors& Errors Fixed (\%) & Precision(\%) & Recall(\%)& F1(\%) & Accuracy(\%) & Time(s) \\
				    \midrule
				FarsiVerb + BM \footnote{Bounded Morphemes} & 15683 & 62.36 & 96.16 & 99.89 &97.99&96.06& 1281 \\
				Parsivar +  FarsiVerb &10554 & 74.63 &99.09  &98.63  &98.64  &97.30 & 1289 \\
				Hazm  + FarsiVerb  & 9159 & 78.01& 97.79 & 99.90 &  98.84&97.70 & 1279 \\
				Hazm + BM  & 8715 & 78.85 & 97.85 & \textbf{99.96} & 98.89&97.81& \textbf{5.75} \\
				Hazm + BM + FarsiVerb  & \textbf{8097}& \textbf{80.56} & \textbf{98.07} & 99.89 &\textbf{98.97}&\textbf{97.96}& 1296 \\

				    \bottomrule

			\end{tabular}
			
		\end{center}
		\end{minipage}
	\end{table}

	\section{Conclusion}
This paper provided a novel work focusing solely on tokenization as a text preprocessing stage. It first discussed tokenization in Persian texts, explained the previous work in Persian tokenizers then depicted the main problems of tokenization in Persian. It discussed all the available open-source tokenizers and how they work and proposed an algorithm to evaluate them. According to the result that is shown in Table \ref{tk_errors_1}, it can be clearly seen that Persian text tokenization needs a lot of improvement. Most of the mistakes in the best version were complex words. Hazm has shown the best performance on Persian Dependency\cite{dataset} among all of the available tokenizers.
Furthermore, it showed that using word bound morphemes and Farsi Verb tokenizer can slightly improve the results. In all of the experiments, it could be seen that irregular complex words are one of the prominent sources of error. Therefore, as future works, by focusing on the ideas of this tokenizer and adding language-specific customization, such as adding a specialized dictionary of irregular complex words, we can achieve a higher level of performance in the task.


\bibliographystyle{ACM-Reference-Format}
\bibliography{main}

\end{document}